\pdfoutput=1

\documentclass[11pt]{article}

\usepackage[]{emnlp2021}

\usepackage{times}
\usepackage{latexsym}

\usepackage[T1]{fontenc}

\usepackage[utf8]{inputenc}
\usepackage{microtype}

\usepackage{amssymb}
\usepackage{algorithm,algorithmic}
\usepackage{booktabs}
\usepackage{multirow}
\usepackage{tabularx}
\usepackage{amsmath}
\usepackage{graphicx}

\usepackage{hyperref}

\makeatletter
\newcommand{\printfnsymbol}[1]{%
  \textsuperscript{\@fnsymbol{#1}}%
}
\makeatother

\title{QEMind: Alibaba's Submission to the WMT21 Quality Estimation Shared Task}

\author{Jiayi Wang\thanks{~~ indicates equal contribution.}  , Ke Wang\printfnsymbol{1}, Boxing Chen, Yu Zhao, {\bf Weihua Luo}, {\bf Yuqi Zhang}\thanks{~~ indicates corresponding author.} \\
Alibaba Group, Hangzhou, China\\
\texttt{\{joanne.wjy,moyu.wk,boxing.cbx\}@alibaba-inc.com},\\ 
\texttt{kongyu@taobao.com}, \texttt{\{weihua.luowh,chenwei.zyq\}@alibaba-inc.com}
}

\begin{document}
\maketitle
\begin{abstract}
Quality Estimation, as a crucial step of quality control for machine translation, has been explored for years. The goal is to investigate automatic methods for estimating the quality of machine translation results without reference translations. In this year's WMT QE shared task, we utilize the large-scale XLM-Roberta pre-trained model and additionally propose several useful features to evaluate the uncertainty of the translations to build our QE system, named \textit{QEMind}. The system has been applied to the sentence-level scoring task of Direct Assessment and the binary score prediction task of Critical Error Detection. In this paper, we present our submissions to the WMT 2021 QE shared task and an extensive set of experimental results have shown us that our multilingual systems outperform the best system in the Direct Assessment QE task of WMT 2020.
\end{abstract}

\section{Introduction}
Quality estimation (QE) aims to predict the quality of a machine translation (MT) system's output without any access to ground-truth translation references or human intervention \cite{blatz2004confidence, specia2009estimating, specia2018findings}. Automatic methods for QE are highly appreciated in MT applications when we expect to efficiently obtain the quality indications for a larget amount of machine translation outputs in a short time, or even at run-time. This paper describes Alibaba's submissions to the WMT 2021 Quality Estimation Shared Task. We developed a novel QE system, called QEMind, that have been applied to two tasks this year, the sentence-level direct assessment (DA) and binary score prediction of Critical Error Detection (CED). 
\begin{figure}[ht]
  \includegraphics[width=\linewidth]{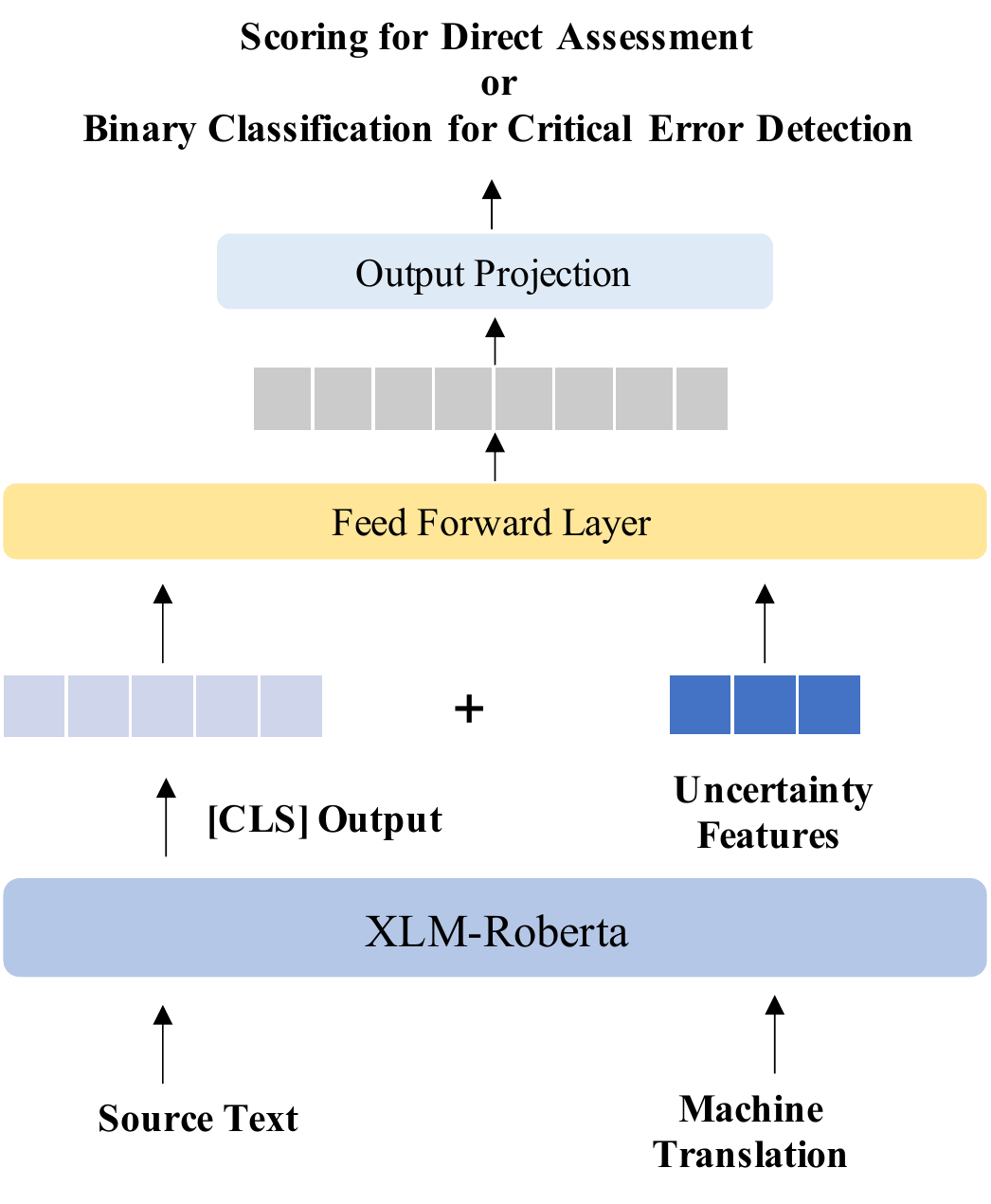}
  \caption{Structure of the uncertainty quantification feature-enhanced model.}
  \label{fig:model}
\end{figure}

Common approaches in the previous years heavily focus on human-crafted rule-based feature engineering mode such as QuEst++ \cite{specia2015multi}. The features extracted are usually fed into traditional machine learning algorithms such as a support vector regression for the sentence-level scoring or a sequence-labeling model with conditional random fields for the word-level labeling respectively. With the development of neural networks applied in machine translation and other NLP tasks, a neural predictor-estimator framework for QE was proposed and achieved better results in WMT 2017 and WMT 2018 QE shared tasks \cite{fan2019bilingual, kim2017predictor}. This framework extensively requires a pre-training procedure with a large amount of parallel corpora in the predictor mode and stacks a downstream estimator mode with additional layers for a supervised regression or classification task. Since 2019, state-of-the-art (SOTA) QE systems \cite{kepler2019unbabel,transquest} have hit record high with transfer learning by leveraging SOTA pre-trained NLP neural network models, for example, mBERT \cite{mbert} and XLM-Roberta \cite{xlm-r}. Till then, only "black-box" QE methods had been mainly used in WMT QE shared tasks.

Furthermore, with the accessibility to the NMT systems, some "glass-box" QE features have been explored and verified to bring improvements upon "black-box" approaches \cite{ist-unbabel}. In addition, \citet{unsupervisedQE} have showed that useful information that are extracted from the MT systems performs good correlation with human judgements of quality. Inspired by these works, we propose more useful features in this paper, among which, some are derived from the NMT systems and others are created via utilizing the masked language model of XLM. We develop our QE systems by incorporating all the features that can potentially evaluate the uncertainty of the machine translations into a supervised QE model based on the transfer learning from XLM-Roberta. We evaluate our method on the Direct Assessment QE tasks of WMT 2020 and WMT 2021 and our experiment results demonstrate the efficiency and versatility of the features we have proposed on the quality estimation in different language pairs.

\section{Task \& Data Set}
We participate the sentence-level Direct Assessment task and Critical Error Detection tasks of this year's QE shared task. (1) For the DA task, we merge 7000 and 1000 labeled data in the training and development data sets as our training set and treat the test20 data set as our development set for each of the seven language pairs. However, for the four zero-shot language pairs, we only have the blind test sets. (2) For the CED task, we observed that the distributions of two classes, NOT and ERR, are extremely unbalanced for all four language pairs. Therefore, we simply up-sample the samples with ERR labels to get a relatively balanced training set. This strategy of data augmentation has also been empirically verified to be valid.
\section{Methodology}
\label{method}
In this section, we provide a complete view of our uncertainty feature enhanced approach, including: 

(1) The overall framework of QEMind is carried out in Section \ref{method:framework}: how uncertainty features are combined with a pre-trained multilingual language model to enhance transfer learning;

(2) Uncertainty features used in QEMind are described in Section \ref{method:feature}: how uncertainty features are defined and extracted for translation quality estimation;

(3) Strategies we applied in the WMT QE shared task to further improve the system's performance, such as data augmentation and model ensemble, are explained in Section \ref{method:strategy}.

\subsection{QEMind Framework}
\label{method:framework}
QEMind follows the general transfer learning procedure while allowing extra meta features to enhance the model. 
We concatenate the source text and machine translation and feed them into the pre-trained XLM-Roberta model to get the output representation of the special [CLS] token. 
Afterwards, the output representation is combined with the normalized uncertainty features described in Section \ref{method:feature}. They are fed into a simple linear regression/classification layer to predict the continuous or binary quality score. 
The architecture of our feature enhanced model is shown in Figure \ref{fig:model}. This model is equivalent to TransQuest's \cite{transquest} when no extra feature is used. 
Considering the size of the training set is small, we have not added extra parameters, such as bottle-neck adapter layers used in \citet{ist-unbabel}, to fuse uncertainty features and the output from XLM-Roberta. 

\subsection{Uncertainty Features}
\label{method:feature}
\citet{unsupervisedQE} proposed several "glass-box" features extracted from the NMT model. Estimating translation quality with these features achieves state-of-the-art results as an unsupervised approach. 
However, the performances of this approach are still far below those of the supervised model from transfer learning \cite{transquest}. 
\citet{ist-unbabel} combined limited "glass-box" features with the hidden state of a bottle-neck adapter layer attached on the output from the XLM-Robert, and the results indicate that these features can bring slight but significant improvements to the transfer learning model. \citet{wang2021glassbox} proposed more unsupervised "glass-box" and "black-box" QE features and investigated further on the contributions of each one to the QE model's performance via a feature-enhanced supervised model. 

Inspired by their work, we explored deeply in the aspect of uncertainty quantification to obtain uncertainty features in this section to enhance the transfer learning model. First, we extend "glass-box" features in \citet{unsupervisedQE} to the \textit{Decoding Probability Features} and the \textit{Monte Carlo Dropout Features}. And then, the \textit{Noised Data Features} are proposed similar to the \textit{Monte Carlo Dropout Features}.

\textbf{Decoding Probability Features.} For auto-regressive sequence generating models like Transformers \cite{transformer}, the decoding probability at each step can be extracted from the softmax layer directly in a "glass-box" setting:
\begin{equation}
    P_{step}^{(\mathbf{x},t,\theta)}=\log P(y_t|\mathbf{y}_{<t},\mathbf{x},\theta)
\end{equation}
where $\mathbf{x}$ represents the input source text and $\mathbf{y}$ is the output machine translation.
$P_{step}$ is a probability sequence with the same length of the generated sequence $\mathbf{y}$. Three statistical indicators of $P_{step}$ can be used to estimate the uncertainty of the output: expectation, standard deviation, and the combined ratio of them:
\begin{equation}
    \mathbb{E}(P_{step}|\mathbf{x},\theta)=\frac{1}{T}\sum_{t=1}^{T}{P_{step}^{(\mathbf{x},t,\theta)}}
\end{equation}
\begin{equation}
\begin{aligned}
    & \sigma(P_{step}|\mathbf{x},\theta)\\
    =& \sqrt{\mathbb{E}(P_{step}^2|\mathbf{x},\theta)-\mathbb{E}^2(P_{step}|\mathbf{x},\theta)}
\end{aligned}
\end{equation}
\begin{equation}
    Combo(P_{step}|\mathbf{x},\theta)=\frac{\mathbb{E}(P_{step}|\mathbf{x},\theta)}{\sigma(P_{step}|\mathbf{x},\theta)}
\end{equation}
Intuitively, larger expectation, smaller deviation and larger combined ratio of $P_{step}$ indicate lower uncertainty and higher quality. $P_{step}$ is an extended version of the $TP$ feature in \citet{unsupervisedQE} and the expectation of $P_{step}$ is the same as $TP$.

\textbf{Monte Carlo Dropout Features.} Monte Carlo (MC) Dropout sampling, that has been exploited in \citet{gal2016dropout}, is an efficient "glass-box" approach to estimate uncertainty. 
It enables random dropout on neural networks during inference and the predictive probabilities through different sampling paths are used to obtain measures of uncertainty \cite{unsupervisedQE}. The output sequences $\hat{\mathbf{y}}$ sampled across stochastic forward-passes by MC dropout with different sampled model parameters $\hat{\theta}$ can be different as well. 
If $\mathbf{y}$ is a high-quality output with low uncertainty, the Monte Carlo sampled outputs $\hat{\mathbf{y}}$ should be close to $\mathbf{y}$ and the variance of $\hat{\mathbf{y}}$ should be low. 
Hence, two measurements of sampling based on text similarity are carried out here:
\begin{equation}
    MC\textit{-}Sim=Sim(\mathbf{y}, \hat{\mathbf{y}}_i)
\end{equation}
\begin{equation}
    MC\textit{-}Sim\textit{-}Inner=\frac{1}{N}\sum_{j=1}^{N}{Sim(\hat{\mathbf{y}}_i,\hat{\mathbf{y}}_j)}
\end{equation}
where $\hat{\mathbf{y}}_i$ is the $i$-th sample of $\hat{\mathbf{y}}$, and $1 \leq i \leq N $.
For the similarity score function, as in \citet{unsupervisedQE}, Meteor metric \cite{meteor} is applied. 
Besides, as a sentence-level probability score, $\mathbb{E}(P_{step})$ can also be calculated with different model parameters $\hat{\theta}$ by MC dropout sampling:
\begin{equation}
    MC\textit{-}P_{step}=\mathbb{E}(P_{step}|\mathbf{x},\hat{\theta})
\end{equation}
The expectation, standard deviation, and combined ratio of $MC\textit{-}Sim$, $MC\textit{-}Sim\textit{-}Inner$ and $MC\textit{-}P_{step}$ are calculated over all MC dropout samples and will be used as "glass-box" uncertainty features. 
Among them, $\mathbb{E}(MC\textit{-}P_{step})$, $\sigma(MC\textit{-}P_{step})$, $Combo(MC\textit{-}P_{step})$, and $\mathbb{E}(MC\textit{-}Sim\textit{-}Inner)$ are equivalent to $D\textit{-}TP$, $D\textit{-}Var$, $D\textit{-}Combo$, and $D\textit{-}Lex\textit{-}Sim$ in \citet{unsupervisedQE}

\textbf{Noised Data Features.} Monte Carlo Dropout approaches mentioned above can be regarded as a robustness test of the NMT model. 
Due to its validity in \citet{unsupervisedQE}, it is rational to believe that a similar way with appropriate noise in the input of MT may perform comparably. 
Therefore, we define the following uncertainty features similar to $MC\textit{-}Sim$, $MC\textit{-}Sim\textit{-}Inner$ and $MC\textit{-}P_{step}$. 
The differences are: (1) the NMT model weights are fixed $\theta$ without MC dropout sampling; (2) the model decodes translations $\tilde{\mathbf{y}}$ with a noised input $\tilde{\mathbf{x}}$.

\begin{equation}
\label{noise_sim}
    Noise\textit{-}Sim=Sim(\mathbf{y}, \tilde{\mathbf{y}}_i)
\end{equation}
\begin{equation}
\label{noise_dsim}
    Noise\textit{-}Sim\textit{-}Inner=\frac{1}{N}\sum_{j=1}^{N}{Sim(\tilde{\mathbf{y}}_i,\tilde{\mathbf{y}}_j)}
\end{equation}
\begin{equation}
\label{noise_tp}
    Noise\textit{-}P_{step}=\mathbb{E}(P_{step}|\tilde{\mathbf{x}},\theta)
\end{equation}
One crucial point in designing this type of features is how to generate noised input $\tilde{\mathbf{x}}$. One solution is a "black-box" way that takes the advantage of the masking strategy of the pre-trained XLM-Roberta. 
Basically, we can mask some words in the source text and get a noised source text by the predictions from the pre-trained model in the masked positions. 
This simple approach only conducts substitutions on $\mathbf{x}$ with the [mask] token, but it limits the diversity of the noised sample inputs.
To enrich the variety of $\mathbf{x}$, we adjust the imitation learning algorithm in \citet{catape} to a simplified version to obtain noised input ${\tilde{\mathbf{x}}}$. 
We "post-edit" the input $\mathbf{x}$ by randomly deleting tokens and inserting masks for several rounds to get $\mathbf{x}_{mask}$. 
Then, the pre-trained XLM-R is used as a masked language model to predict the tokens in the masked positions of $\mathbf{x}_{mask}$ to get the post-edited $\tilde{\mathbf{x}}$. 
Pseudo codes of this "post-editing" algorithm is provided in Algorithm \ref{alg:noise}. 

\begin{algorithm}[t]
\caption{Generate Noise Input with "Post-Editing"}
\begin{algorithmic}[1]
\label{alg:noise}
\REQUIRE input $\mathbf{x}=\{x_t|t=1,2,...,T\}$, hyper-parameters $R$, $p_i$, $p_d$.
\STATE Initialize $\mathbf{x}_{mask}=\mathbf{x}$
\FOR{$r=1,...,R$} 
	\STATE $\mathbf{x}_{mask}$ = randomly delete tokens from $\mathbf{x}_{mask}$ with probability $p_d$
	\STATE $\mathbf{x}_{mask}$ = randomly insert special \textit{<mask>} tokens into $\mathbf{x}_{mask}$ with probability $p_i$
\ENDFOR
\STATE $\tilde{\mathbf{x}}=MLM(\mathbf{x}_{mask})$, where $MLM$ is a pre-trained masked language model.
\RETURN $\tilde{\mathbf{x}}$
\end{algorithmic} 
\end{algorithm}

\subsection{Strategies}
\label{method:strategy}
\textbf{Multilingual Training.} Considering zero-shot language pairs in the DA task, we mix up all seven language pairs' training data to fine-tune the XLM-Roberta model and predict on the whole test set including zero-shot language pairs. We have tried two different ways of mixing up training data from different language pairs to fine-tune XLM-Roberta: (1) source sentence + translation sentence; (2) English sentence + non-English sentence. Our experimental results demonstrate that multilingual models usually perform better than bilingual models trained on a single language pair, but there is no prominent difference in performance of the two 
different multilingual strategies. We keep both multilingual models and bilingual models for model ensemble.

\textbf{Data Augmentation.} Two data augmentation strategies are applied for the CED task. First, considering the imbalance between positive and negative samples in the CED dataset, we up-sample the data with\textit{ERR} labels in each language pair to obtain a balanced dataset. Secondly, inspired by examples provided by the organizer, we have also tried to replace the original machine translation with a back-translated sentence and hope that the gap between the source sentence and the back-translated sentence can provide insights of the detection of potential critical errors. The back translations come from the released ML50 multilingual translation model \cite{ml50}.

\textbf{Model Ensemble.} For the DA task, models trained with different multilingual strategies and different uncertainty features are ensembled by averaging predicted scores. 
While for the CED task, we average classification probability outputs from models trained with different data augmentation strategies and uncertainty features to obtain ensemble results. 
We apply a greedy ensemble strategy. 
First, all models are sorted by their performance on the development sets. 
Then, upon the best single model, we take one more model into the ensemble at each step until there is no more performance gain on the development sets or the maximum step is reached. We set the maximum step to avoid overfitting on the development sets.
\begin{table*}[]
\centering
\begin{tabular}{@{}lccccccc@{}}
\toprule
\multicolumn{1}{c}{\multirow{2}{*}{Model}} & \multicolumn{2}{c}{High-Resource} & \multicolumn{3}{c}{Mid-Resource}                    & \multicolumn{2}{c}{Low-Resource}  \\
\multicolumn{1}{c}{}                       & En-De           & En-Zh           & Et-En           & Ro-En           & Ru-En           & Si-En           & Ne-En           \\ \midrule
OpenKiwi (Official Baseline)               & 0.1455          & 0.1902          & 0.4770          & 0.6845          & 0.5459          & 0.3737          & 0.3860          \\
TransQuest Single                          & 0.4669          & 0.4779          & 0.7748          & 0.8982          & 0.7734          & 0.6525          & 0.7914          \\
QEMind-Bi                          & 0.4463          & 0.4471          & 0.7569          & 0.8961          & 0.7990          & 0.6443          & 0.7988          \\
QEMind-Multi                       & 0.5107          & 0.4762          & 0.8031          & 0.9009          & 0.7984          & 0.6775          & 0.7958          \\
QEMind-Multi + UNC & \textbf{0.5746} & \textbf{0.5094} & \textbf{0.8156} & \textbf{0.9039} & \textbf{0.8044} & \textbf{0.6843} & \textbf{0.8130} \\ \midrule
TransQuest Ensemble                        & 0.5539          & 0.5373          & 0.8240          & 0.9082          & 0.8082          & 0.6849          & 0.8222          \\
QEMind Ensemble                            & \textbf{0.6054} & \textbf{0.5445} & \textbf{0.8410} & \textbf{0.9173} & \textbf{0.8273} & \textbf{0.7079} & \textbf{0.8374} \\ \bottomrule
\end{tabular}
\caption{The Pearson's correlation between model predictions and human DA judges on the WMT 2020 QE test sets}
\label{tab:da20}
\end{table*}
\begin{table*}[htp]
\centering
\begin{tabular}{@{}lccccccc@{}}
\toprule
\multicolumn{1}{c}{\multirow{2}{*}{Model}} & \multicolumn{2}{c}{High-Resource} & \multicolumn{3}{c}{Mid-Resource} & \multicolumn{2}{c}{Low-Resource} \\
                       & En-De           & En-Zh           & Et-En     & Ro-En     & Ru-En    & Si-En           & Ne-En          \\ \midrule
Official Baseline      & 0.4025          & 0.5248          & 0.6601    & 0.8175    & 0.6766   & 0.5127          & 0.7376         \\
QEMind Single    & 0.5281          & 0.5635          & 0.7909    & 0.8954    & 0.7893   & 0.5769          & 0.8406         \\
QEMind Ensemble  & \textbf{0.5666}          & \textbf{0.6025}          & \textbf{0.8117}    & \textbf{0.9082}    & \textbf{0.8060}   & \textbf{0.5956}          & \textbf{0.8667}         \\ \bottomrule
\end{tabular}
\caption{Pearson's correlations results of 2021 DA task on non-zero-shot language pairs}
\label{tab:da21}
\end{table*}

\begin{table}[]
\setlength{\tabcolsep}{2pt}
\begin{tabular}{@{}lcccc@{}}
\toprule
\multicolumn{1}{c}{Model}                & En-Ja  & En-Cs  & Km-En  & Ps-En  \\ \midrule
Official Baseline     & 0.2301 & 0.3518 & 0.5623 & 0.4760 \\
QEMind Single   & 0.3354 & 0.5456 & 0.6509 & 0.6159 \\
QEMind Ensemble & \textbf{0.3589} & \textbf{0.5816} & \textbf{0.6787} & \textbf{0.6474} \\ \bottomrule
\end{tabular}
\caption{Pearson's correlations results of 2021 DA task on zero-shot language pairs}
\label{tab:da21zero}
\end{table}
\begin{table}[]
\setlength{\tabcolsep}{2pt}
\begin{tabular}{@{}lcccc@{}}
\toprule
\multicolumn{1}{c}{Model} & En-Cs           & En-De           & En-Ja           & En-Zh           \\ \midrule
QEMind                    & 0.3915          & 0.4629          & 0.2559          & 0.2629          \\
QEMind + BK               & \textbf{0.4257} & \textbf{0.4914} & 0.2471          & 0.2800          \\
QEMind + UNC              & 0.4111          & 0.4859          & \textbf{0.2606} & \textbf{0.2897} \\ \midrule
QEMind Ensemble           & \textbf{0.4864} & \textbf{0.5257} & \textbf{0.3325} & \textbf{0.3587} \\ \bottomrule
\end{tabular}
\caption{Matthews correlation results of WMT 2021 CED task on development sets}
\label{tab:cedev}
\end{table}

\begin{table}[]
\setlength{\tabcolsep}{2pt}
\begin{tabular}{@{}lcccc@{}}
\toprule
\multicolumn{1}{c}{Model} & En-Cs           & En-De           & En-Ja           & En-Zh           \\ \midrule
Official Baseline         & 0.3875          & 0.3974          & 0.2139          & 0.1873          \\
QEMind-Single             & 0.4129          & 0.4257          & 0.2139          & 0.2356          \\
QEMind Ensemble           & \textbf{0.4539} & \textbf{0.4797} & \textbf{0.2601} & \textbf{0.2777} \\ \bottomrule
\end{tabular}
\caption{Matthews correlation results of WMT 2021 CED task on test sets}
\label{tab:cetest}
\end{table}
\section{Experiments}
\subsection{Model Settings}
We follow the model settings of Transquest \cite{transquest} to fine-tune our QE model based on the XLM-Roberta large model with a classification/regression head on a single P100 GPU. The training batch size is set to 8 and the training process takes about 2 hours to convergence. For the DA task, the total number of parameters of QEMind with uncertainty features is 560981507; if no uncertainty features are used, it is 560941571. And for the CED task, the numbers of parameters with and without uncertainty features are 560982532 and 560942596 respectively.

\subsection{Experiments of DA task}
We conduct all experiments and evaluate our model on last year's test sets to optimize model configurations for each language pair. 
In particular, the model performed best on all seven language pairs in average is selected to generate submissions for zero-shot language pairs. 

The Pearson's correlations between our model's predictions and the human DA judges (z-standardized mean DA score) are shown in Table \ref{tab:da20}. 
\textit{TransQuest Single} and \textit{TransQuest Ensemble} are the best single and ensemble models of \citet{transquest}, which is the winner system of last year's DA task. 
\textit{QEMind-Bi} and \textit{QEMind-Multi} are models without uncertainty features, between which, the difference is that the model is trained on bilingual data or mixed multilingual data. 
\textit{QEMind-Multi + UNC} is the complete QEMind model enhanced by various uncertainty features described in Section \ref{method:feature}. 
Finally, predictions from bilingual models, multilingual models, and uncertainty features enhanced models are ensembled following Section \ref{method:strategy}, marked as \textit{QEMind Ensemble} in the table.  

Results on the DA test sets of WMT 2020 show that:
(1) multilingual strategies work well on this task, especially for high-resource language pairs; 
(2) the uncertainty features enhanced multilingual model achieves the highest performance among all single models, which verifies that these uncertainty features are useful to all language pairs and can be fused in multilingual models.
(3) ensemble of multiple models of different settings can further improve the performance of QEMind systems.

We pick the best single and ensemble models for each language pair and produce predictions on the newly released blind test sets of WMT 2021, including the 4 zero-shot language pairs. Results of Pearson's correlations are shown in Table \ref{tab:da21} and Table \ref{tab:da21zero}.

\subsection{Experiments of CED task}

We test different strategies and uncertainty features on the CED development sets. Brief results of Matthews correlations (MCC) on development sets are shown in Table \ref{tab:cedev}. 
All models are trained on up-sampled training data of each language pair. 
From the results observations, compared to \textit{QEMind}, which only applies up-sampling on the training data, the strategies of back-translation (\textit{QEMind + BK}) and uncertainty features (\textit{QEMind + UNC}) can achieve comparable or better performances. The ensemble of all these models makes a significant improvement. Similar to the DA task, the best single and ensemble models are picked to generate our final submissions. Results on test sets of this year are listed in Table \ref{tab:cetest}.

\section{Conclusion}
This paper introduces our machine translation quality estimation model, QEMind, for the sentence-level Direct Assessment and Critical Error Detection tasks of WMT 2021. We propose novel features to estimate the uncertainty of machine translations and incorporate them into the transfer learning from the large-scale pre-trained model, XLM-Roberta. Besides, three important strategies are particularly utilized for improving the QE system's performance such as multilingual training, data augmentation and model ensemble. Our system has achieved the first ranking in average Pearson correlation across all languages, including the zero-shot ones in the multilingual DA task of WMT 2021. 
\section*{Acknowledgements}
This work is supported by National Key R\&D Program of China (2018YFB1403202).
\bibliography{anthology,custom}
\bibliographystyle{acl_natbib}




\end{document}